\documentclass[letterpaper]{article}

\usepackage[T1]{fontenc}

\usepackage{geometry}
\usepackage{setspace}

\usepackage[
  backend=biber,
  style=chem-acs,
  articletitle=true,
  pageranges=true,
  maxbibnames=10,
  minbibnames=10,
  giveninits=true,
  doi=false,
  url=true,
  isbn=false,
  eprint=true
]{biblatex}

\usepackage{xpatch}

\DeclareFieldFormat[inproceedings]{title}{#1}

\xpatchbibdriver{inproceedings}
  {\usebibmacro{byauthor}%
   \setunit{\addspace}%
   \usebibmacro{in:}}
  {\usebibmacro{title}%
   \printunit{\adddot\addspace}%
   \usebibmacro{byauthor}%
   \setunit{\addspace}%
   \usebibmacro{in:}}
  {}
  {\PackageWarning{jpclrefs}{Could not patch the inproceedings bibliography driver}}

\usepackage{graphicx}
\usepackage{float}
\newfloat{scheme}{htbp}{los}
\floatname{scheme}{Scheme}
\floatname{chart}{Chart}
\newfloat{graph}{htbp}{loh}

\usepackage{chemformula} 
\usepackage[version = 4]{mhchem} 

\usepackage{lipsum}
\usepackage{graphicx}
\usepackage{amsmath}
\usepackage{amsfonts}

\usepackage{algorithm}        
\usepackage{algorithmicx}     
\usepackage{algpseudocode}

\usepackage[final]{changes}
 
\usepackage{authblk}
\author[1]{Pengfei Liu}
\author[2]{Shuang Ge}
\author[3]{Xiaobo Wang}
\author[3]{Xin Liu}
\author[4]{Jun Tao}
\author[5]{Yan Li}
\author[3,6]{Chao Liu}
\author[1]{Ling Chen}
\author[7]{Zhixiang Ren*}

\affil[1]{Guangzhou Key Laboratory of Forensic Multi-Omics for Precision Identification, Southern Medical University, Guangzhou, Guangdong, 510515, China}
\affil[2]{Tsinghua Shenzhen International Graduate School, Tsinghua University, Shenzhen, Guangdong 518000, China}
\affil[3]{Guangdong Provincial Key Laboratory of Psychoactive Substances Monitoring and Safety, Anti-Drug Technology Center of Guangdong Province (National Anti-Drug Laboratory Guangdong Regional Center), Guangzhou, 510230, China}
\affil[4]{School of Computer Science and Engineering, Sun Yat-Sen University, Guangzhou, Guangdong 510006, China}
\affil[5]{Department of Nuclear Medicine, Xiyuan Hospital, China Academy of Chinese Medical Sciences, Beijing 100091, China}
\affil[6]{State Key Laboratory of Bioactive Molecules and Druggability Assessment, Guangdong Basic Research Center of Excellence for Natural Bioactive Molecules and Discovery of Innovative Drugs, Jinan University, Guangzhou 510632, China}
\affil[7]{Shanghai Smart Logic Technology Co., Ltd., Shanghai 200443, China}

\title{A Multi-task Large Reasoning Model for Molecular Science}
\date{*Email: jason.zhixiang.ren@outlook.com}

\begin{document}

\maketitle

\begin{abstract}
Artificial intelligence in molecular science must move beyond pattern recognition toward chemically valid and interpretable reasoning.
We present a task-adaptive large reasoning model that integrates chemical knowledge through a synergistic multispecialist architecture, chain-of-thought supervision, and molecule-informed reinforcement learning.
Task-conditioned routing coordinates prediction and inference specialists across 10 molecular tasks spanning molecular description and generation, nomenclature translation, property prediction, and reaction prediction.
The model outperforms more than 20 general-purpose and molecular large language models, improves aggregate performance over the base model by 50.3\%, and surpasses the leading molecular multitask baseline on most tasks.
Analyses of specialist representations and reasoning pathways reveal task-specific adaptation while retaining interpretable chemical inference.
A case study further demonstrates an integrated workflow for central nervous system candidate generation, property screening, molecular interpretation, and retrosynthetic planning.
These results \deleted{establish a general framework for knowledge-guided molecular reasoning and design}\added{demonstrate a versatile multi-task framework for knowledge-guided molecular reasoning and design, with the potential to serve as a core task engine for future molecular science agents}.
\end{abstract}

\section*{}

\begin{figure}[t!]
    \centering
    \includegraphics[width=\textwidth]{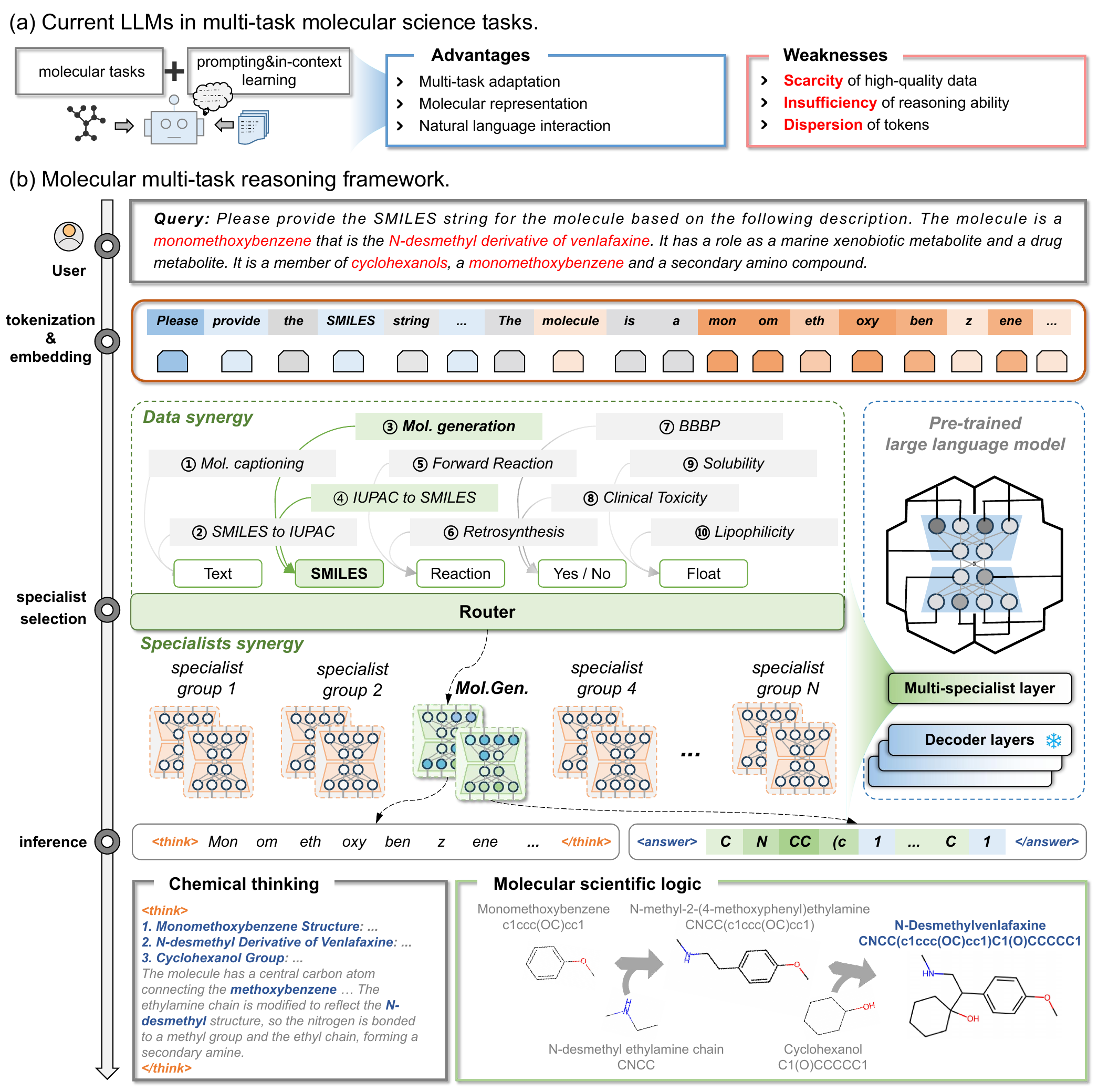}
    \caption{
    \textbf{Overview of the reasoning framework.}
    (a) Current LLMs in multi-task molecular science tasks, showcasing their core capabilities and inherent challenges.
    (b) Molecular multi-task reasoning framework, detailing the process from user query to inference, featuring tokenization and embedding, specialist selection via a router, and a multi-specialist layer within a pre-trained LLM.
    This framework unifies diverse molecular tasks through data synergy and embeds chemical logic into CoT reasoning for science-grounded outputs, with an example showcasing a text-based molecular generation task.}
    \label{fig:multispecialist_framework}
\end{figure}

Artificial intelligence (AI) has rapidly become a driving force in molecular science, with significant applications in drug discovery \cite{jimenez2020drug}, materials design \cite{huang2021artificial}, and chemical synthesis planning \cite{zhang2025large}.
Traditional experimental methods and quantum chemical calculations, while foundational, are often time-consuming, expensive, and limited by their inability to handle large-scale molecular datasets efficiently.
Conventional machine learning approaches have improved prediction accuracy for small datasets but struggle with scalability, feature engineering complexity, and capturing intricate molecular interactions \cite{wigh2022review}.
In particular, deep learning techniques, such as graph neural networks (GNNs), have been widely applied to model molecular structures as graphs, enabling tasks like property prediction and reaction forecasting \cite{li2025kolmogorov}.
These methods excel in capturing local structural features but often suffer from high computational costs, limited scalability to large datasets, and challenges in modeling long-range dependencies within complex molecular sequences.

In contrast, the Transformer architecture \cite{vaswani2017attention} addresses these limitations through its self-attention mechanism, which efficiently processes sequential data like SMILES \cite{weininger1988smiles} strings, facilitating better generalization and interpretability in molecular representations.
The advent of Transformer-based models \cite{devlin2019bert}\cite{radford2019language}\cite{raffel2020exploring} has strengthened the understanding and generation capabilities of sequential data.
The emergence of ChatGPT \cite{schulman2022chatgpt} has sparked a surge in LLMs, inspiring their adaptation in molecular science.
For instance, MolT5 \cite{edwards2022translation} enables translation between molecules and natural language, which supports tasks such as molecule captioning and property description.
BioT5+ \cite{pei2024biot5+} extends this paradigm to biological contexts by integrating IUPAC \cite{favre2013nomenclature} names for enhanced understanding in drug discovery.
GIT-Mol \cite{liu2024git} and MoleculeSTM \cite{liu2023multi} further incorporate multi-modal inputs for comprehensive molecular analysis.
Those methods have demonstrated strong predictive power across diverse molecular tasks, enabling advances in property prediction \cite{zheng2025large}, de novo molecular design, reaction prediction \cite{liu2025self}, and molecule captioning.
Nevertheless, the majority of existing approaches are still primarily data-driven, focusing on statistical correlations \cite{kuang2024impact} rather than capturing the underlying scientific principles \cite{liu2025quantitative}.
They are typically \textbf{designed for single tasks}, limiting their generalizability to diverse multi-task scenarios.
Moreover, they often \textbf{lack explicit reasoning mechanisms}, which necessitate logical inference and seamless integration of domain-specific knowledge.

Recent advances in multi-task molecular LLMs have begun to address these gaps by scaling foundation models for broader applicability.
For example, LLaSMol \cite{yullasmol}, ChemLLM \cite{zhang2024chemllm}, and InstructBioMol \cite{zhuang2025advancing} have achieved excellent performance in multi-task and multi-modal molecular learning, highlighting the potential of scaling molecular foundation models.
However, these models primarily serve as high-performing predictors, lacking robust interpretability and explicit reasoning capabilities, which limits their application in scientific discovery scenarios that require transparent decision-making.
While models such as o1 \cite{openai2024openaio1card} and DeepSeek-R1 \cite{deepseekai2025deepseekr1}, among other advanced large-scale reasoning models, have introduced advanced reasoning frameworks, they also adapt to Mixture-of-Experts (MoE) \cite{shazeer2017outrageously} architectures, enabling flexibility across diverse tasks.
However, these models encounter challenges with domain-specific accuracy and scalability when applied to molecular contexts.
DeepMind’s TxGemma model \cite{wang2025txgemma} combines conversational capabilities with molecular multi-task learning, though this integration results in decreased task performance.
These limitations highlight a critical need for a multi-task molecular reasoning model that achieves high predictive accuracy and possesses robust reasoning capabilities.

Since molecular science relies on established chemical rules, structured knowledge, and logical reasoning, there is a growing need for models that integrate such knowledge to deliver accurate predictions and science-based reasoning.
To tackle these challenges, we introduce a synergistic multi-specialist knowledge reasoning model that embeds molecular logic into a task-adaptive framework, as illustrated in Figure~\ref{fig:multispecialist_framework}.
Unlike prediction-focused models, our approach incorporates chemical knowledge via chain-of-thought (CoT) reasoning \cite{wei2022chain}.
We construct an instruction dataset comprising 93K instances to train prediction specialists, focusing on molecular knowledge representation, and a dataset with chemical CoT reasoning, embedding molecular science constraints to align AI decisions with scientific principles, consisting of 3.5K high-quality instances for inference specialists.
In contrast to MoE architectures, we leverage \textbf{data synergy} (joint training of data with similar knowledge while isolating disparate knowledge) and \textbf{specialist synergy} (collaboration between prediction specialists excelling in molecular representation and inference specialists proficient in structured reasoning) to enhance molecular knowledge representation and reasoning.

Enhanced by molecule-informed reinforcement learning, this approach aligns reasoning and answers with chemical accuracy through sparse CoT data, boosting data efficiency and performance.
Our model outperforms over 20 multi-task LLMs across 10 molecular tasks, achieving improvements of up to 10\% compared to the state-of-the-art baseline (LLaSMol).
To illustrate its practical utility, we present a case study on CNS drug candidate design, where the model streamlines the workflow from text-based molecule generation to retrosynthetic planning, enabling rapid analysis of blood-brain barrier penetration and toxicity profiles.
It can accelerate molecular design cycles, reducing the time from conceptual sketches to viable candidates from weeks to hours while embedding interpretable chemical reasoning throughout.
This capability fosters reliable scientific exploration by bridging empirical data and science-driven insights, supporting innovations in drug discovery and materials design.


\section*{}

\label{sec:result}


\textbf{Molecular multi-task reasoning framework.}
We follow the experimental design of the representative molecular multi-task model LLaSMol, selecting ten representative molecular science tasks to assess our model’s capabilities.
These tasks encompass text-generation challenges, including molecule captioning and SMILES generation \cite{edwards2022translation} for creating textual descriptions and molecular structures, respectively, as well as SMILES-to-IUPAC and IUPAC-to-SMILES translations for interconverting chemical notations.
Additionally, the benchmark includes property prediction \cite{wu2018moleculenet} with classification tasks, such as blood-brain barrier penetration (BBBP) and clinical toxicity (ClinTox), as well as regression tasks like water solubility (ESOL) and lipophilicity.
It also contains forward reaction prediction and retrosynthesis, focusing on reaction outcome forecasting and the design of synthetic pathways.
These tasks serve as key evaluations for LLMs in molecular science, as they are ideally suited for reasoning-oriented assessments due to their reliance on logical inference and domain-specific knowledge.

\textit{Knowledge-infused dataset design.}
We employed a stratified sampling method based on molecular feature distributions to construct a 93K instruction-following dataset from over 2 million data points, training prediction specialists to enhance molecular knowledge representation while ensuring efficient model performance.
Additionally, we conducted deep sampling to build a 3.5K high-quality dataset, completed with CoT annotation, to train inference specialists focused on molecular knowledge inference.
Further details are available in Supplementary Information A.

\textit{Synergistic multi-specialist architecture.}
The model integrates multiple specialist modules, coordinated by a task-conditioned hard-routing mechanism, to tackle diverse molecular tasks.
It harnesses data synergy through the joint training of data containing similar knowledge and the isolation of disparate knowledge.
This amplifies related knowledge, with specialist assignments tailored to task knowledge types.
For instance, text-based SMILES generation and IUPAC-to-SMILES are handled by the same specialist. Forward reaction and retrosynthesis prediction are managed by a different specialist. For property prediction, BBBP and ClinTox are assigned to separate experts despite both being classification tasks. Their distinct feature distributions optimize performance.
Moreover, it leverages specialist synergy through collaboration between prediction specialists excelling in molecular representation and inference specialists proficient in structured reasoning.
Prediction specialists and inference specialists work together as pairs to jointly execute reasoning tasks.
This makes reasoning more directed, avoiding errors caused by overly long reasoning chains.

\textit{Knowledge-guided alignment strategy.}
Prediction specialists and inference specialists undergo initial independent training.
As reasoning length increases, inference specialists may deviate from the initial answers provided by prediction specialists.
To address this challenge, we implement reinforcement learning with task-specific molecular science reward functions.
For instance, the SMILES generation reward integrates SMILES validity and similarity to the target molecule.
This approach enhances the consistency of reasoning between prediction specialists and inference specialists.

\begin{figure}[t!]
\centering
\includegraphics[width=\textwidth]{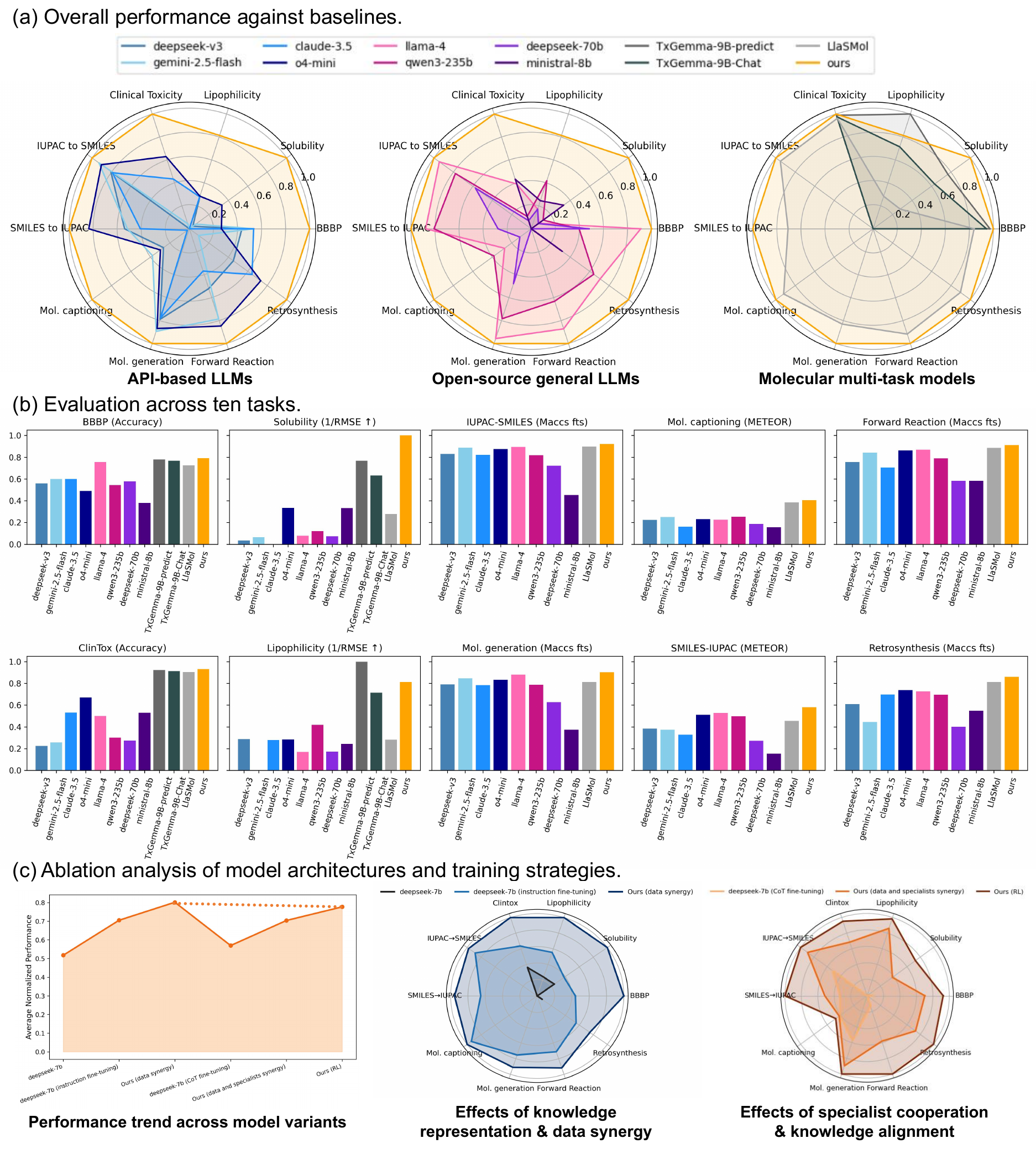}
\caption{
\textbf{Comprehensive evaluation of model performance and synergy.}
(a) Overall performance against baselines, comparing our model with over ten representative LLMs across core metrics as detailed in (b), with all radar plot metrics normalized.
(b) Detailed metric values across the ten tasks.
(c) Ablation analysis of model architectures and training strategies, contrasting experimental setups to validate the efficacy of data synergy and specialist synergy.}
\label{fig:results}
\end{figure}

\textbf{Performance evaluation and ablation studies.}
We evaluate the effectiveness of our model by assessing its performance across ten representative molecular science tasks.
These assessments follow established evaluation protocols as outlined in \cite{wu2018moleculenet, yullasmol}.
Each task is comprehensively evaluated using five or more assessment metrics, with representative metrics carefully selected to address task-specific challenges.
METEOR scores \cite{banerjee2005meteor} measure the quality of text-generation tasks by evaluating semantic similarity between generated and reference descriptions.
MACCS molecular fingerprint \cite{durant2002reoptimization} similarity gauges structural accuracy for SMILES generation and reaction prediction tasks by comparing molecular fingerprints.
\added{Chemical validity was evaluated separately from fingerprint similarity.
The generated SMILES were parsed, canonicalized, and sanitized using rdkit; the output was deemed chemically valid.
For reaction prediction, all predicted reactants and products were subjected to the same structure-level validation before the reaction metrics were calculated.}
Accuracy is crucial for the success of classification tasks, such as BBBP and ClinTox, as it ensures the correct identification of molecular properties.
For regression tasks like ESOL and Lipophilicity, we use 1/RMSE, which serves as an inverse measure of prediction error.
The complete experimental setups, evaluation metrics, and results are comprehensively detailed in Supplementary Information B.

\textit{Baselines.}
To benchmark our model, we compare it against over 20 baselines, selecting representative models from three distinct series, as illustrated in Figure~\ref{fig:results} (a), which highlights a subset of competitive models for clarity.
From API-based LLMs, we include DeepSeek-V3 \cite{liu2024deepseek}, Gemini-2.5-Flash \cite{comanici2025gemini}, Claude-3.5 \cite{anthropic2025claude35}, and o4-mini \cite{openai2025o4mini}, chosen for their advanced prompt-guided capabilities.
Open-source general LLMs are represented by Qwen3-235B \cite{yang2025qwen3}, DeepSeek-70B \cite{deepseekai2025deepseekr1}, MiniStral-8B \cite{mistral2025ministrale8b}, and LLaMA-4 \cite{meta2025llama4}, selected for their broad adaptability and community validation.
For molecular multi-task models, we opt for TxGemma-9B-predict, TxGemma-9B-Chat, and LLaSMol \cite{yullasmol}, reflecting specialized designs in this domain.
The TxGemma series \cite{wang2025txgemma} lacks training for text generation, molecular generation, and reaction prediction tasks. To ensure fairness, we exclude these tasks from comparison with TxGemma.
API-based LLMs and open-source general LLMs depend on prompt-guided responses, while molecular multi-task models are locally deployed and reproduced for consistent testing.

\textit{Overall performance insights.}
Notably, our model outperforms all baselines in overall performance, achieving superior results in all tasks except Lipophilicity.
As depicted in Figure~\ref{fig:results} (a), the three radar charts clearly illustrate our model’s significant advantage over other models.
API-based LLMs tend to generate more hallucinations, with longer reasoning processes increasingly deviating from accurate answers or knowledge, leading to reduced accuracy.
In contrast, LLaMA-4 exhibits moderate molecular knowledge and capability in sequence/text generation, but underperforms in quantitative tasks such as property prediction.
Knowledge-enriched Molecular multi-task models, including ours, demonstrably outperform API-based LLMs and Open-source general LLMs, reflecting the benefit of domain-specific expertise.
However, our model falls slightly behind in the Lipophilicity regression task, potentially due to its specialized training on regression-focused datasets that our model has not fully optimized for.
Compared to other methods, LLaSMol achieves a solid performance in multi-molecular task accuracy, as shown in Figure~\ref{fig:results} (b).
Relative to the state-of-the-art molecular multi-task model LLaSMol, our task metrics demonstrate an improvement of nearly 6\%.

\begin{figure}[t!]
\centering
\includegraphics[width=\textwidth]{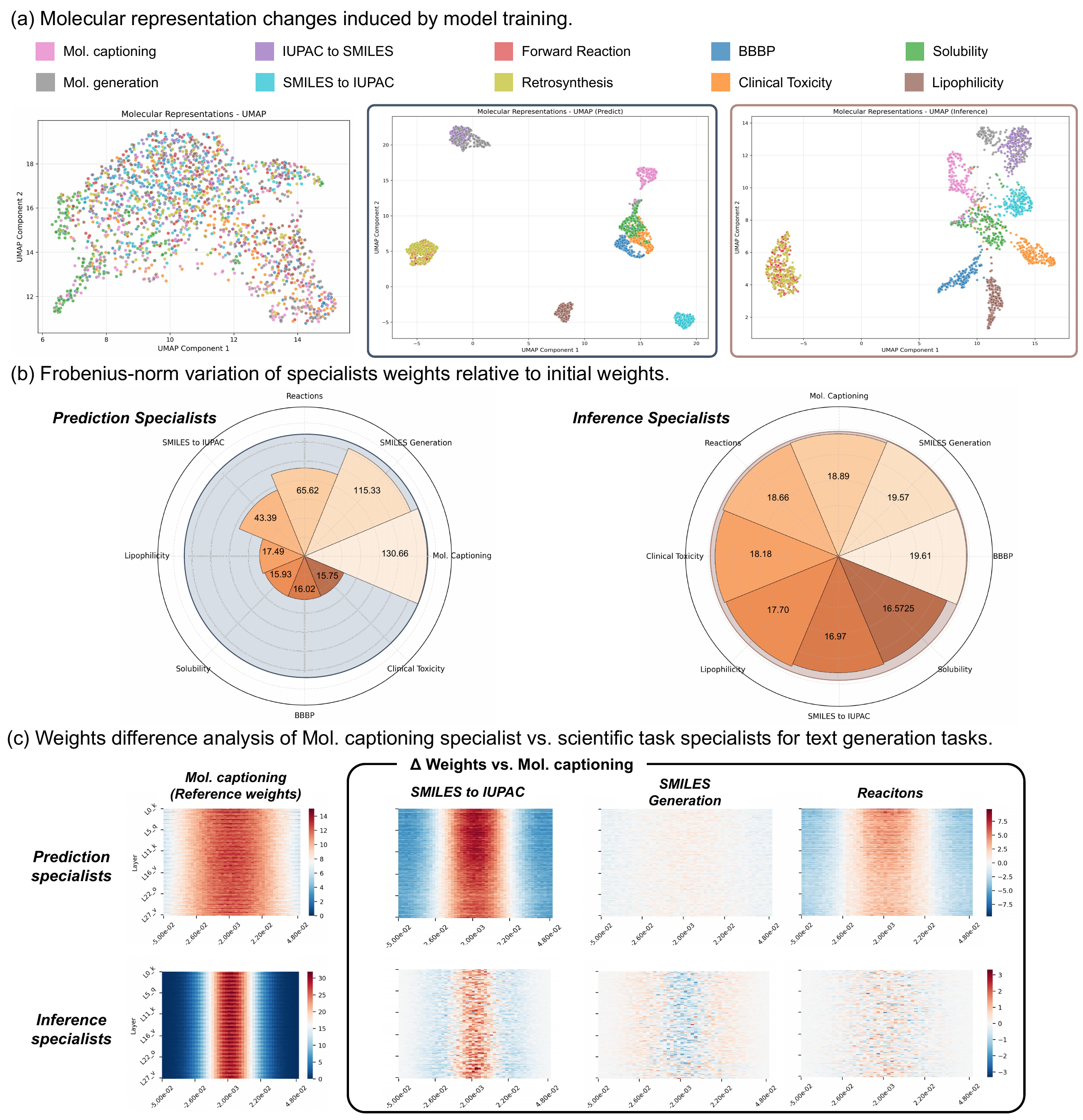}
\caption{\textbf{Post-training adaptations in specialist modules.}
(a) Molecular representation changes induced by model training, visualized via UMAP dimensionality reduction on sampled molecules from ten tasks processed through each specialist.
(b) Frobenius norm variation of specialists' weights relative to initial weights, quantifying adaptation degrees.
(c) Weight difference analysis between the molecule captioning specialist and scientific task specialists for text generation tasks.
Heatmaps show 2D histograms of merged LoRA A/B weights per layer, with y-axis as sorted layers (e.g., Layer\_n\_q), x-axis as binned weights [-0.05, 0.05], and cell intensities as probability densities (darker = higher density, estimated via 50-bin histograms with units of 1/weight\_value).
Left: absolute density for captioning specialist (linear norm); right: signed density differences $\Delta\rho$ (symmetric norm).}
\label{fig:specialist_analysis}
\end{figure}

\textit{Ablation study findings.}
To quantify the benefits of our core strategies---\textbf{data synergy} and \textbf{specialist synergy}---we conduct ablation studies using DeepSeek-7B as the base model, with the mean performance across tasks serving as the aggregate evaluation metric.
As shown in Figure~\ref{fig:results} (c), instruction fine-tuning enhances molecular task capabilities, and subsequent specialist synergy further boosts reaction prediction and SMILES generation, elevating the aggregate metric from 0.705 to 0.801.
In contrast, CoT-only fine-tuning on DeepSeek-7B yields modest gains (0.517 to 0.569) due to the smaller dataset size and increased inference length.
Applying data and specialist synergy to this variant then improves performance to 0.704.
Finally, the integration of reinforcement learning, incorporating task-specific reward functions to guide reasoning, raises the metric to 0.777 (a 50.3\% improvement over 0.517).
Starting from a baseline model of DeepSeek-7B fine-tuned on the instruction dataset with an initial metric of 0.705, this enhancement achieves an improvement of over 10\%, realizing a multi-task molecular reasoning model that delivers high predictive accuracy and robust reasoning capabilities.

\textbf{Specialist evolution analysis.}
To gain deeper insights into the internal dynamics of our model, we analyze the evolution of specialist modules before and after training, focusing on changes in molecular representation capabilities and internal weight distributions.
This examination reveals how data synergy and specialist synergy foster distinct adaptations, enabling the model to balance knowledge representation (via predict specialists) and logical inference (via inference specialists) across molecular tasks.

As illustrated in Figure~\ref{fig:specialist_analysis} (a), pre- and post-training Uniform Manifold Approximation and Projection (UMAP) \cite{mcinnes2018umap} projections vividly demonstrate the evolution of task-specific embeddings developed by specialists.
This demonstrates unique molecular knowledge interpretations in the latent vector space for both prediction specialists and inference specialists.
Prediction specialists cluster representations more tightly for structural fidelity.
Inference specialists expand clusters to accommodate reasoning pathways. Figure~\ref{fig:specialist_analysis} (b) quantifies whether the overall magnitude of the specialist weights increases after training. For each weight matrix $W$, we use the Frobenius norm:
\begin{equation}
\|W\|_F
=
\left(
\sum_{i,j}w_{i,j}^2
\right)^{1/2}.
\end{equation}
The signed change in weight magnitude is then calculated as:
\begin{equation}
\Delta M
=
\|W_{\mathrm{trained}}\|_F
-
\|W_{\mathrm{initial}}\|_F.
\end{equation}
A positive value indicates an increase in the overall weight magnitude, whereas a negative value indicates a decrease.
Prediction specialists undergo more aggressive fine-tuning to capture domain-specific patterns, which reflect substantial updates for enhanced knowledge representation.
Inference specialists prioritize stability for consistent CoT application.
They show moderate changes, indicating refined integration of reasoning logic.

The weight difference heatmap in Figure~\ref{fig:specialist_analysis} (c) underscores the granularity of specialization.
The molecule captioning specialist serves as the reference due to its foundational role, as all task reasoning processes can be fundamentally transformed into captioning tasks.
In contrast, the compared text-generation tasks further embody greater scientific rigor.
As illustrated in Figure~\ref{fig:specialist_analysis} (c), the weight difference heatmap reveals distinct activation patterns between prediction specialists and inference specialists across these text-generation tasks.
To quantify these differences, we compute per-layer histograms of flattened LoRA weights $w$ from matrices A and B, yielding probability densities $\rho_l(b)$ for layer $l$ and bin $b$ via normalized histograms:
\begin{equation}
\rho_l(b) = \frac{n_l(b)}{N_l \Delta b},
\end{equation}
where $n_l(b)$ is the count in bin $b$, $N_l$ is total weights in layer $l$, and $\Delta b=0.002$ is bin width (yielding $\rho$ in units of 1/weight\_value). The base distribution uses linear normalization $\rho_{\text{captioning}} \in [0, \max(\rho_{\text{captioning}})]$. Differences are
\begin{equation}
\Delta \rho_l(b) = \rho_l(b) - \rho_{\text{captioning}}(b),
\end{equation}
symmetrically normalized over $[\min(\Delta \rho), \max(\Delta \rho)]$ across all layers and specialists, revealing task-specific deviations such as broader early-layer spreads in SMILES-to-IUPAC, mid-layer centralization in SMILES generation, and late-layer sparsity in reaction prediction.
This highlights our framework's adaptability to multi-task scenarios by leveraging specialized routing.
Such differentiation validates the efficacy of the design and supports enhanced performance in diverse molecular reasoning applications.

\begin{figure}[t!]
\centering
\includegraphics[width=0.85\textwidth]{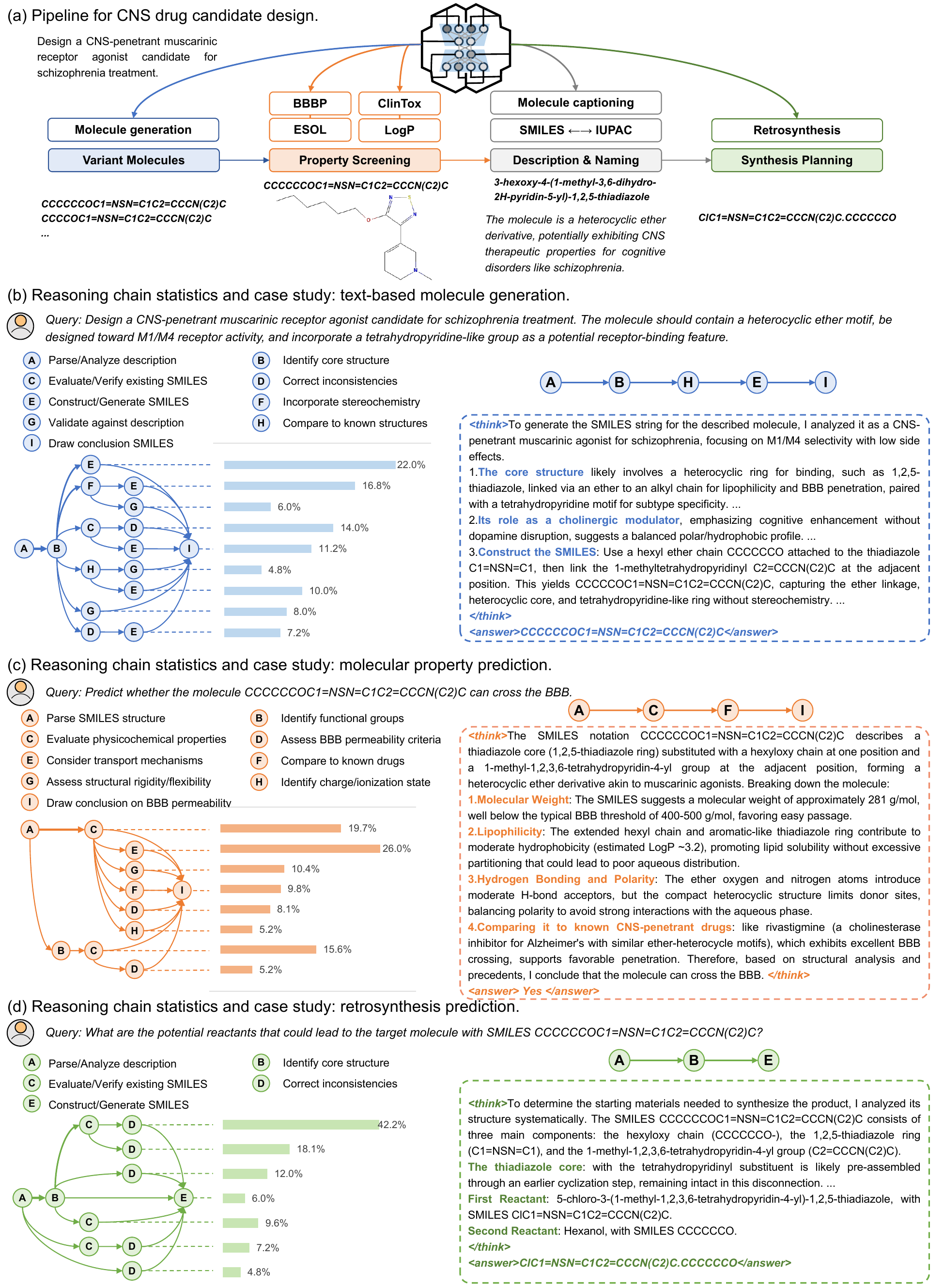}
\caption{\textbf{Case study: the pipeline for CNS drug candidate development.}
(a) Our model integrates multi-task reasoning for CNS drug screening and analysis.
(b) Case: Text-based molecule generation for designing a CNS-penetrant muscarinic receptor agonist candidate.
(c) Case: Molecular property prediction, exemplified by the BBBP task.
Here, the A$\to$C$\to$I chain accounts for 19.7\% and A$\to$C$\to$E$\to$I for 26.0\% of reasoning paths, illustrating distribution via bar charts of multi-step inference flows.
(d) Case: Retrosynthesis prediction, identifying potential reactants for the target molecule.}
\label{fig:cases}
\end{figure}

\textbf{\deleted{Case study for CNS drug candidate design.}\added{Proof-of-concept workflow for CNS candidate evaluation.}}
\added{The model's capabilities are particularly well-suited for CNS drug candidate design, where interpretable CoT reasoning and specialist synergy address critical challenges, such as BBBP and toxicity profiling.}
As shown in Figure~\ref{fig:cases}, \deleted{our model serves as a versatile framework for constructing diverse molecular screening pipelines, enabling seamless integration of multi-task reasoning across property prediction, generation, and synthesis to accelerate candidate evaluation}\added{we use a case study to demonstrate how molecular generation, property screening, molecular interpretation, and retrosynthetic planning can be connected within a unified workflow}.
\added{This case study is intended to illustrate the coordination of multiple molecular tasks rather than to claim the discovery of a genuinely new therapeutic candidate.}
\deleted{The model's capabilities are particularly well-suited for CNS drug candidate design, where interpretable CoT reasoning and specialist synergy address critical challenges, such as BBBP and toxicity profiling.}
\deleted{In this case study, we focus on xanomeline (SMILES: CCCCCCOC1=NSN=C1C2=CCCN(C2)C) as an exemplar.}
\added{In this case study, the model was prompted with the desired pharmacological objective and structural features, without being provided with the name or SMILES of xanomeline.
The resulting structure, CCCCCCOC1=NSN=C1C2=CCCN(C2)C, corresponds to xanomeline.}
Xanomeline is a muscarinic receptor agonist and the centrally acting component of Cobenfy, a combination of xanomeline and trospium chloride approved by the Food and Drug Administration in September 2024 for the treatment of schizophrenia in adults \cite{paul2024muscarinic}.
It marks the first such therapy to target muscarinic receptors (M1 and M4 subtypes) rather than dopamine receptors, offering a novel mechanism for cognitive symptom relief with reduced extrapyramidal side effects.

The pipeline begins with text-based molecule generation to produce candidates, followed by property prediction and screening (BBBP, ESOL, Lipophilicity, and ClinTox) to \deleted{select viable candidates}\added{prioritize structures predicted to exhibit property profiles compatible with further CNS-oriented evaluation}.
Mechanism description is achieved via molecule captioning and SMILES-to-IUPAC conversion for interpretability.
The pipeline concludes with the prediction of feasible synthetic routes through retrosynthesis.
\added{For the generated structures, the predicted BBB penetration, toxicity, solubility, and lipophilicity should be interpreted as computational estimates rather than experimental evidence.
Moreover, for additional generated variants, structural features specified in the generation prompts may indicate potential compatibility with muscarinic receptor activity but do not establish receptor binding, agonistic activity, safety, or therapeutic efficacy.}

\textbf{Reasoning chain distribution analysis.}
We assess the reasoning capabilities of our model by analyzing the distribution of reasoning chains across its outputs.
This study focuses on three representative tasks: text-based molecule generation, molecular property prediction (BBBP), and retrosynthesis.
We examine the diversity of reasoning elements and their proportional contributions.
Figure~\ref{fig:cases} provides an overview of these distributions, highlighting the model's adaptability to varied molecular reasoning demands, with further details in Supplementary Information C.

The analysis reveals distinct reasoning patterns tailored to each task. Text-based molecule generation emphasizes structural construction, while BBBP prediction focuses on comparative property assessment, and retrosynthesis prioritizes structural disassembly.
These variations highlight the model's capacity to adjust its reasoning process, driven by the synergy between prediction specialists and inference specialists.
This adaptability enhances overall performance across diverse molecular tasks, as validated by consistent output accuracy.
The case studies demonstrate the practical impact of these reasoning chains.
Each task employs a distinct sequence of logical steps, yielding high-fidelity outputs, including validated SMILES strings, accurate permeability predictions, and feasible reactant pairs.

\subsection*{}

\textbf{Molecular multi-task models.}
Our model \deleted{offers distinct advantages over existing molecular multi-task models by integrating domain-specific reasoning and achieving superior performance across a wide range of molecular tasks}\added{provides a unified framework for integrating domain-specific reasoning across a broad range of molecular tasks}.
\added{A unified model is advantageous for workflows requiring the coordinated execution of molecular generation, property prediction, structural interpretation, and retrosynthetic analysis, as it enables shared molecular representations and reduces the need to deploy separate models. However, task-specific models may remain preferable when the primary objective is to optimize a single endpoint, because different molecular tasks rely on distinct chemical information and learning objectives, and joint training may introduce performance trade-offs.}

\added{The individual technical components of the framework, including specialist routing, CoT supervision, low-rank adaptation, and reinforcement learning.
The main contribution of this study lies in their coordinated integration across chemically distinct tasks, enabling molecular representation, generation, property prediction, and reaction analysis within a unified framework.}
Unlike traditional approaches that rely heavily on data-driven predictions, our framework leverages data and specialist synergy to enhance adaptability, distinguishing itself from conventional MoE models through task-tailored specialist collaboration, as evidenced by its outperformance of baseline LLMs.
However, a limitation arises from the computational overhead of managing multiple specialists, which may hinder scalability for extremely large datasets.
Future research could focus on optimizing resource allocation and exploring continuous learning strategies to adapt dynamically to evolving datasets.

\textbf{\added{Chemical scope and applicability.}}
\added{The present framework is primarily evaluated on SMILES-based benchmark tasks and therefore represents two-dimensional molecular connectivity more directly than stereochemistry or three-dimensional conformational effects.
Reaction predictions also do not explicitly account for solvents, catalysts, temperature, yields, or competing pathways, while benchmark labels may contain assay-dependent variability and experimental uncertainty.
Consequently, the model outputs should be interpreted as computational predictions within the evaluated task space; stereochemical assignment, conformational analysis, reaction feasibility, and biological activity still require external calculations or experimental validation.}

\textbf{Knowledge-integrated reasoning models.}
Building on this multi-task foundation, our approach excels in embedding chemical knowledge into reasoning processes, setting it apart from knowledge-integrated models that often lack dynamic adaptability.
By employing CoT reasoning and reinforcement learning, the model generates interpretable outputs, bridging the gap between data-driven and science-grounded paradigms.
A notable limitation is the dependency on high-quality CoT datasets, which are currently limited in size and scope, potentially restricting generalization.
Looking ahead, we aim to enrich these datasets through automated knowledge extraction and iterative refinement, broadening the model’s versatility across diverse molecular scenarios.

\textbf{Interpretability.}
The strength of our model lies in its ability to deliver transparent reasoning chains through CoT processes, thereby enhancing trustworthiness in molecular science with decision-making insights grounded in scientific principles.
This clarity distinguishes it from opaque black-box models, boosting reliability for applications such as drug discovery.
Yet, the complexity of certain molecular tasks can lead to overly detailed reasoning chains, potentially confusing users or masking key insights.
To overcome this, future research may explore interactive visualization interfaces or condensed reasoning formats, thereby strengthening the model’s practical dependability in scientific settings.

\textbf{LLM-based molecular science agents.}
Leveraging the multi-task and reasoning capabilities of our model, it serves as a solid framework for advancing LLM-based molecular science agents.
Its capacity to blend specialist expertise and adapt to varied demands establishes it as a flexible core for intelligent agents.
For instance, Google DeepMind's AI Co-Scientist, a multi-agent system powered by Gemini 2.0, acts as a virtual collaborator for generating hypotheses and research proposals \cite{gottweis2025aicoscientist}.
Similarly, Sakana AI's ``The AI Scientist'' exemplifies how such agents can automate complex research processes through an automated pipeline for end-to-end scientific paper generation and discovery \cite{lu2024aiscientistfullyautomated}.
These agents are adept at navigating complex workflows, employing reasoning akin to that of human experts.
To overcome the limitation of static knowledge integration identified earlier, future developments could incorporate automated tool invocation and data querying.
Additionally, real-time feedback mechanisms and interactive training could be integrated.
This paves the way for fully autonomous molecular science assistants. Moreover, it drives innovation in fields like drug discovery and materials design.

In summary, this study introduces \deleted{a reasoning model that redefines molecular science through enhanced reasoning and superior multi-task performance}\added{a versatile multi-task molecular reasoning model that integrates chemical representation, generation, property prediction, and reaction analysis within a unified framework}.
By integrating chemical knowledge via CoT reasoning, leveraging data and specialist synergy, and employing reinforcement learning, the model outperforms baseline LLMs across 10 molecular tasks, demonstrating significant gains in accuracy and interpretability.
\deleted{These advancements pave the way for a new paradigm in molecular research, fostering intelligent design solutions beyond traditional data-driven approaches.}
\added{These capabilities support the use of the model as a general molecular task engine and provide a practical foundation for future molecular science agents that coordinate prediction, design, analysis, and external chemical tools.}
Future directions include optimizing computational efficiency, expanding high-quality CoT datasets, and developing real-time interactive agents, promising further breakthroughs in the field.

\section*{Computational Methods}

\label{sec:met}

\begin{figure}[t!]
\centering
\includegraphics[width=0.9\textwidth]{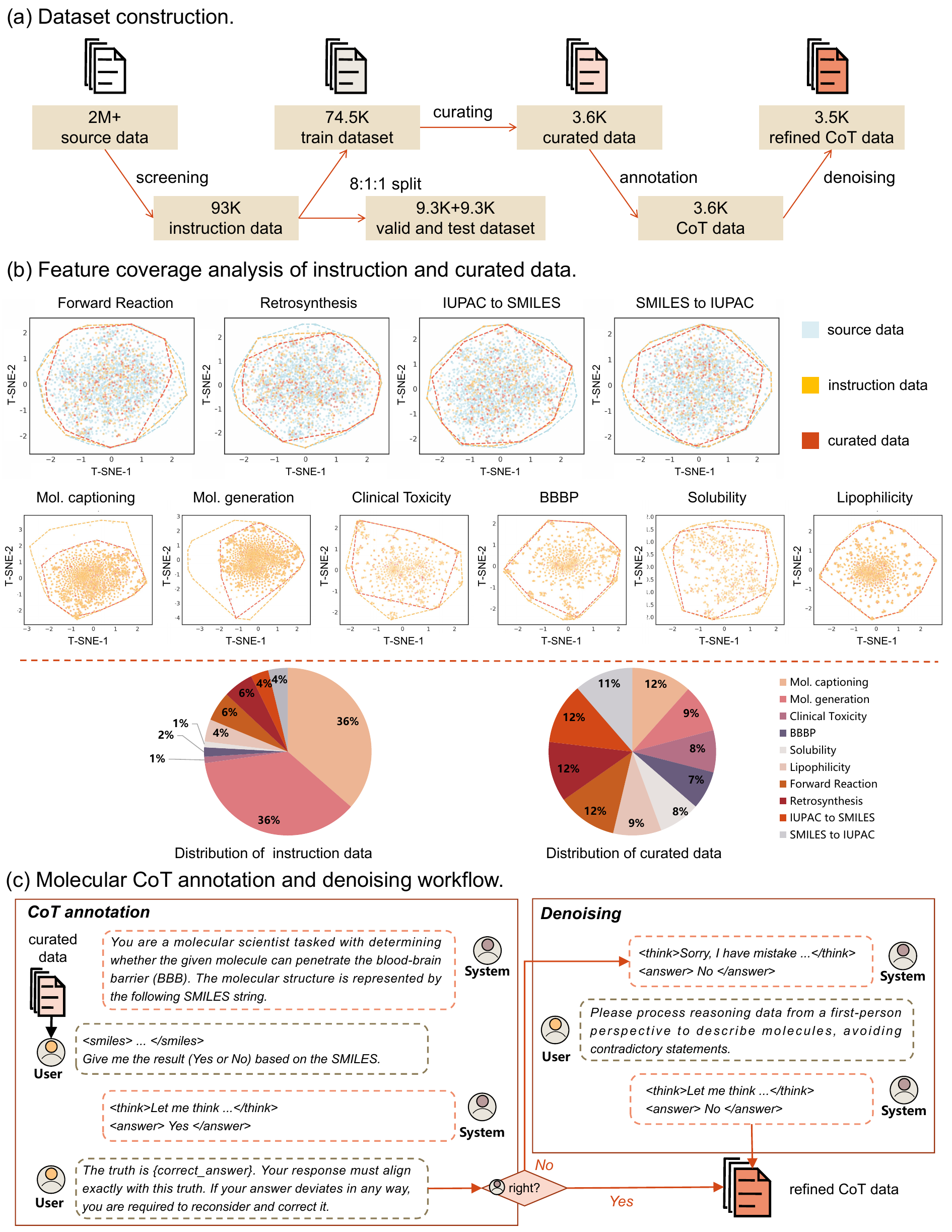}
\caption{
\textbf{Dataset construction and CoT data refinement.}
(a) Construction and splitting of the 93K instruction dataset, together with selection of the curated subset.
(b) t-SNE visualization of feature coverage across the source, instruction, and curated datasets.
(c) CoT annotation and denoising, resulting in approximately 3.5K refined CoT instances.
}
\label{fig:dataset_construction}
\end{figure}

\textbf{Dataset construction.}
The dataset is sourced from a variety of tasks aligned with the LLaSMol framework \cite{yullasmol}, encompassing 10 molecular science challenges. These include property prediction tasks such as BBBP, ESOL (Solubility), Lipophilicity, and ClinTox, drawn from the MoleculeNet benchmark \cite{wu2018moleculenet}, alongside molecule captioning and generation derived from ChEBI-20 \cite{edwards2022translation}.
Additionally, IUPAC-to-SMILES and SMILES-to-IUPAC translations, as well as forward and retrosynthesis reaction predictions, are sampled from LLaSMol.
This diverse task set ensures comprehensive coverage of molecular reasoning and prediction capabilities, forming the foundation for our model’s multi-task performance.
The construction of the 93K instruction dataset and 3.6K curated dataset involves a structured process, as depicted in Figure~\ref{fig:dataset_construction} (a).

\textit{Instruction dataset.}   
The 93K instruction dataset is meticulously crafted from an extensive initial pool, diverging from LLaSMol’s approach by refining its voluminous IUPAC-to-SMILES, SMILES-to-IUPAC translations, and reaction prediction data.
While these data offer depth, their scale can lead to diminishing training efficiency beyond a certain threshold.
To counter this, we employ a strategic sampling compression technique, which reduces the dataset size while preserving its quality.
This process, guided by Figure~\ref{fig:dataset_construction} (a), ensures optimal performance.
The dataset is further shaped through a stratified sampling method based on molecular feature distributions, as depicted in Figure~\ref{fig:dataset_construction} (b).
Starting with over 2 million source data points, we curate the 93K instruction dataset, for which the data volume for each remains at or below the ten-thousand level to optimize training efficiency.
This process leverages molecular representations extracted from the DeepSeek-7B model, followed by dimensionality reduction using t-distributed stochastic neighbor embedding (t-SNE) \cite{van2008visualizing} and sampling according to the resulting feature distributions.
The sampling process
can be approximated as
\begin{equation}
S_i = P_i\left(\operatorname{t\text{-}SNE}\left(f_i(R_i)\right)\right),
\label{equ: sample}
\end{equation}
where $S_i$ represents the sampled subset for the $i$-th task,
$f_i(R_i)$ denotes the molecular representations $R_i$ extracted by the DeepSeek-7B model, and $P_i$ denotes the sampling function defined according to the feature distribution in the t-SNE space.
For the remaining six tasks, the corresponding datasets are incorporated directly without sampling, ensuring that all task-specific data is included in the final instruction dataset.
The complete instruction dataset $I$ is formed by taking the union of the task-specific subsets:
\begin{equation}
I = \bigcup_{i=1}^{N} S_i,
\label{eq:dataset_union}
\end{equation}
where $S_i$ denotes the subset corresponding to the $i$-th task and $N$ is the total number of task-specific subsets.
Subsequently, we partition the dataset $I$ into training, validation, and test sets at an 8:1:1 ratio.

\textit{Curated dataset.}  
The 3.6K curated dataset is derived as a high-quality subset from the instruction dataset $ I $, specifically designed to support the subsequent annotation of CoT data.
This dataset undergoes a rigorous selection process, aligning with the stratified sampling approach to maintain feature diversity and task relevance. Its construction complements $ I $ by providing a focused resource for refining inference capabilities, with proportional adjustments mirroring the task distributions.
The data extraction method for the curated dataset is consistent with the approach used for the instruction dataset subsets (see equation \ref{equ: sample}).

\textit{Knowledge annotation and denoising.}  
The annotation and denoising of the Molecular CoT dataset follow a rigorous workflow, illustrated in Figure~\ref{fig:dataset_construction} (c). Initially, we utilize DeepSeek-R1 in a dialogue-driven format to prompt reasoning on our questions, validated against correct answers. Successful query-think-answer triplets are extracted into the CoT dataset.
For incorrect responses, DeepSeek-R1 is prompted to self-correct, though these answers often contain anchor segments (e.g., ``I already know the correct answer is ...''), necessitating denoising. Using DeepSeek-V3, we cleanse contradictory content, with 5\% of the data manually validated to confirm the quality of the resulting 3.5K refined CoT dataset.

\begin{figure}[t!]
\centering
\includegraphics[width=\linewidth]{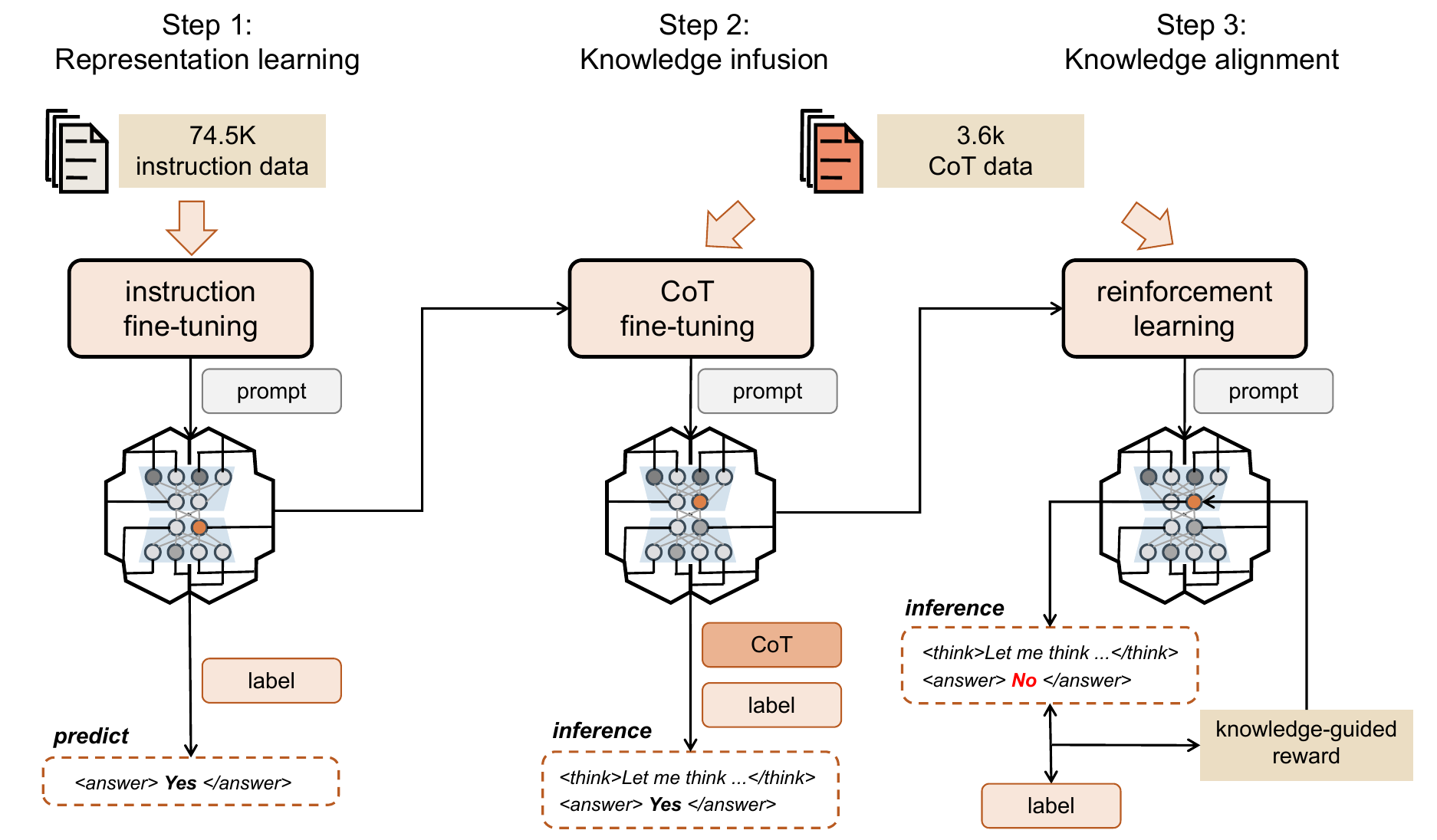}
\caption{Three-step training strategy for the model.
Step 1: Representation learning via instruction fine-tuning on 74.5K instruction data.
Step 2: Knowledge infusion through CoT fine-tuning on 3.6K CoT data.
Step 3: Knowledge alignment using reinforcement learning with knowledge-guided rewards.}
\label{fig:training}
\end{figure}

\textbf{Model.}
Our reasoning model for molecular science is built upon the DeepSeek-7B pre-trained LLM, leveraging a multi-specialist layer integrated within its decoder architecture to handle diverse molecular tasks.
The overall framework combines data synergy across tasks with specialist synergy for enhanced reasoning, as illustrated in Figure~\ref{fig:multispecialist_framework}.
The training strategy follows a three-step process: representation learning via instruction fine-tuning on 74.5K data points, knowledge infusion through CoT fine-tuning on 3.6K samples, and knowledge alignment using reinforcement learning with knowledge-guided rewards, as depicted in Figure~\ref{fig:training}.
This approach embeds chemical logic into the model's inference, enabling task-adaptive reasoning that mimics molecular scientists.

\textit{Model architecture.}
The model architecture extends the standard Transformer decoder with a multi-specialist layer, allowing for efficient handling of multi-task molecular scenarios.
At its core, the model employs the Qwen architecture, a GPT variant, which processes tokenized inputs through self-attention mechanisms. The attention computation for a given input sequence is defined as:
\begin{equation}
\text{Attention}(Q, K, V) = \text{softmax}\left(\frac{QK^T}{\sqrt{d_k}}\right)V,
\end{equation}
where $Q$, $K$, and $V$ are the query, key, and value matrices, respectively, and $d_k$ is the dimension of the keys. This mechanism captures dependencies in molecular representations such as SMILES strings.
To enable efficient adaptation without full parameter tuning, we incorporate Low-Rank Adaptation (LoRA) \cite{hu2022lora} into the Transformer layers.
LoRA approximates weight updates with low-rank matrices, formulated as:
\begin{equation}
W' = W + \Delta W = W + BA,
\end{equation}
where $W$ is the original weight matrix, $B \in \mathbb{R}^{d \times r}$ and $A \in \mathbb{R}^{r \times k}$ are low-rank matrices with rank $r \ll \min(d, k)$, and $\Delta W$ represents the update.

\textit{Multi-specialist layer.}
This layer introduces specialist groups and a task-conditioned router for task-specific processing, promoting data synergy across related tasks and specialist synergy between prediction and inference specialists.
We define eight specialist groups, grouped by output formats and data distributions to realize data synergy (joint training of data containing similar knowledge, such as text-based tasks like molecule captioning and SMILES-to-IUPAC, while isolating disparate knowledge, such as separate groups for classification tasks like BBBP and ClinTox due to distinct feature distributions): (1) Molecule captioning (text output); (2) SMILES to IUPAC (text output, distinct data); (3) Molecular generation and IUPAC to SMILES (SMILES output); (4) Forward reaction prediction and retrosynthesis (reaction/SMILES output); (5) BBBP (Yes/No classification); (6) Clinical Toxicity (Yes/No classification, separate for data variance); (7) Solubility (float regression); (8) Lipophilicity (float regression).

The task-conditioned output is computed as
\begin{equation}
O(q,t)
=
\sum_{i=1}^{N}
r_i(t)\,sg_i(q),
\label{eq:specialist_routing}
\end{equation}
where $q$ is the input query, $t$ is the task identity, $sg_i$ denotes the function of the $i$-th specialist group, and $N=8$ is the number of specialist groups. The binary routing indicator is defined as $r_i(t)=\mathbb{I}[i=g(t)]$, where $g(t)$ maps each task to its corresponding specialist group.
Therefore, $r_i(t)\in\{0,1\}$ and $\sum_{i=1}^{N}r_i(t)=1$, such that one specialist group is activated for each task. This task-conditioned routing strategy enables related tasks to share specialist parameters while maintaining separation between tasks with distinct output formats or data distributions. Within the selected group, the prediction specialist and the corresponding inference specialist collaborate during reasoning-oriented inference.

\textit{Training strategy.}
The training strategy begins with representation learning through instruction fine-tuning on 74.5K multi-task instruction data, unifying diverse molecular tasks under a common prompt format. This step optimizes the cross-entropy loss:
\begin{equation}
\mathcal{L}_{\text{inst}} = -\sum_{t=1}^{T} \log p(y_t | y_{<t}, x),
\end{equation}
where $x$ is the input prompt, $y$ is the target sequence, and $T$ is the sequence length.
Next, knowledge infusion via CoT fine-tuning on 3.6K curated samples embeds chemical reasoning into the model.
At its core, CoT prompting enhances LLMs by decomposing complex problems into intermediate reasoning steps, mimicking human-like logical progression to improve accuracy and interpretability in tasks requiring multi-step inference.
This is particularly suited to molecular science, where it bridges structural analysis to predictive outcomes.
By augmenting inputs with explicit reasoning traces and structuring outputs as $\langle$think$\rangle$ steps followed by $\langle$answer$\rangle$, our approach fosters deliberate, scientist-like cognition.
The loss remains cross-entropy but targets CoT-augmented sequences to align generation with chemical logic.

Finally, knowledge alignment employs the REINFORCE algorithm, a policy gradient method that optimizes the model's generation policy by maximizing expected rewards over reasoning-augmented sequences.
Rewards are first standardized to reduce variance:
\begin{equation}
\hat{r} = \frac{r - \mathbb{E}[r]}{\sigma_r + \epsilon},
\end{equation}
where $\mathbb{E}[r]$ is the mean reward, $\sigma_r$ is the standard deviation, and $\epsilon = 10^{-6}$ is a small constant for numerical stability. The REINFORCE loss is then computed as
\begin{equation}
\mathcal{L}_{\text{RL}} = -\mathbb{E} \left[ \sum_t \log \pi(a_t | s_t) \cdot \hat{r} \right],
\end{equation}
where $\pi(a_t | s_t)$ denotes the log-probability of action $a_t$ (token) given state $s_t$ (context). The overall reward $r$ integrates task performance and reasoning quality:
\begin{equation}
r = \alpha \cdot r_{\text{answer}} + \beta \cdot r_{\text{think}},
\end{equation}
with $\alpha=0.8$, $\beta=0.2$, $r_{\text{answer}}$ task-specific (e.g., accuracy for classification like BBBP/Clintox, inverse RMSE for regression like ESOL/Lipo, BLEU or SMILES similarity for generation/reaction tasks), and $r_{\text{think}}$ assessing reasoning via length (Gaussian centered at 1569 characters) and diversity (unique word ratio).
This guides the model to resolve conflicts and align with molecular scientific logic.

\textit{Hardware and software conditions.}
The hardware configuration includes a single H20-NVLink GPU with 96GB memory, paired with 16 vCPUs from an AMD EPYC 9K84 96-Core Processor, operating on Ubuntu 22.04.
The software environment is built on Python 3.12, with PyTorch 2.5.1 as the core deep learning framework, utilizing CUDA 12.4 for GPU acceleration.
Key libraries include transformers 4.51.3 for the pre-trained LLM, pandas 2.2.3 for data handling, peft 0.15.2 for LoRA implementation, and rdkit 2025.3.2 for molecular structure processing. Optimization is performed using the Adam optimizer with a learning rate of \(1 \times 10^{-4}\), with batch sizes ranging from 4 to 16 for prediction specialist training and 1 to 2 for inference specialist training.

\section*{Data and Software Availability:}
The dataset and software can be accessed at \url{https://github.com/AI-HPC-Research-Team/Mol-Reasoning} for non-commercial use.

\section*{Author Information:}
\subsection*{Corresponding Authors:}
Zhixiang Ren -- Shanghai Smart Logic Technology Co., Ltd., Shanghai 200443, China; \\Email: jason.zhixiang.ren@outlook.com\\

\subsection*{Author Information:}
Pengfei Liu -- Guangzhou Key Laboratory of Forensic Multi-Omics for Precision Identification, Guangzhou, Guangdong, 510515, China\\
Shuang Ge -- Tsinghua Shenzhen International Graduate School, Tsinghua University, Shenzhen, Guangdong 518000, China\\
Xiaobo Wang -- Guangdong Provincial Key Laboratory of Psychoactive Substances Monitoring and Safety, Anti-Drug Technology Center of Guangdong Province (National Anti-Drug Laboratory Guangdong Regional Center), Guangzhou 510230, China\\
Xin Liu -- Guangdong Provincial Key Laboratory of Psychoactive Substances Monitoring and Safety, Anti-Drug Technology Center of Guangdong Province (National Anti-Drug Laboratory Guangdong Regional Center), Guangzhou 510230, China\\
Jun Tao -- School of Computer Science and Engineering, Sun Yat-Sen University, Guangzhou, Guangdong 510006, China\\
Yan Li -- Department of Nuclear Medicine, Xiyuan Hospital, China Academy of Chinese Medical Sciences, Beijing 100091, China\\
Chao Liu -- Guangdong Provincial Key Laboratory of Psychoactive Substances Monitoring and Safety, Anti-Drug Technology Center of Guangdong Province (National Anti-Drug Laboratory Guangdong Regional Center), Guangzhou 510230, China; State Key Laboratory of Bioactive Molecules and Druggability Assessment, Guangdong Basic Research Center of Excellence for Natural Bioactive Molecules and Discovery of Innovative Drugs, Jinan University, Guangzhou 510632, China\\
Ling Chen -- Guangzhou Key Laboratory of Forensic Multi-Omics for Precision Identification, Southern Medical University, Guangzhou, Guangdong, 510515, China

\section*{Author Contributions:}
\begin{itemize}
  \item \textbf{Pengfei Liu}: Conceptualization, methodology, data curation, software, model training, formal analysis, visualization, and writing of the original draft.
  \item \textbf{Shuang Ge}: Formal analysis, validation, and writing--review and editing.
  \item \textbf{Xiaobo Wang}: Chemical data validation, forensic chemistry analysis, and resources.
  \item \textbf{Xin Liu}: Data curation, chemical structure validation, and resources.
  \item \textbf{Jun Tao}: Methodology, supervision, and computational resources.
  \item \textbf{Yan Li}: Clinical interpretation, case study validation, and funding acquisition.
  \item \textbf{Chao Liu}: Chemical validation, supervision, and funding acquisition.
  \item \textbf{Ling Chen}: Supervision, project administration, and funding acquisition.
  \item \textbf{Zhixiang Ren}: Conceptualization, supervision, and project administration.
\end{itemize}

\section*{Declaration of Competing Interest}
The authors declare that they have no known competing financial interests or personal relationships that could have appeared to influence the work reported in this paper.

\section*{Acknowledgment}

This study was supported by the National Key Research and Development Program of China (No.2024YFC3306605), the Science and Technology Program of the Ministry of Public Security of China (No.2023JSM07), the National Natural Science Foundation of China (Grant no. 82371901), and the Basic and Applied Research Program funded by the Basic Research Project of Guangzhou (No. 202201011630).

\section*{Supporting information}
Supplementary information.pdf: Contains comprehensive details on dataset sources, evaluation metrics, comparison models, and detailed results, along with illustrative case studies and CoT reasoning processes to support the main findings.

\printbibliography

@article{jimenez2020drug,
  title = {{Drug Discovery with Explainable Artificial Intelligence}},
  author = {Jim{\'e}nez-Luna, Jos{\'e} and Grisoni, Francesca and Schneider, Gisbert},
  journal = {Nat. Mach. Intell.},
  volume = {2},
  pages = {573--584},
  year = {2020}
}

@article{huang2021artificial,
  title = {{Artificial Intelligence in Materials Modeling and Design}},
  author = {Huang, JS and Liew, JX and Ademiloye, AS and Liew, Kim Meow},
  journal = {Arch. Comput. Methods Eng.},
  volume = {28},
  pages = {3399--3413},
  year = {2021}
}

@article{zhang2025large,
  title = {{Large Language Models to Accelerate Organic Chemistry Synthesis}},
  author = {Zhang, Yu and Han, Yang and Chen, Shuai and Yu, Ruijie and Zhao, Xin and Liu, Xianbin and Zeng, Kaipeng and Yu, Mengdi and Tian, Jidong and Zhu, Feng and others},
  journal = {Nat. Mach. Intell.},
  volume = {7},
  pages = {1010--1022},
  year = {2025},
  doi = {10.1038/s42256-025-01066-y}
}

@article{zheng2025large,
  title = {{Large Language Models for Scientific Discovery in Molecular Property Prediction}},
  author = {Zheng, Yizhen and Koh, Huan Yee and Ju, Jiaxin and Nguyen, Anh TN and May, Lauren T and Webb, Geoffrey I and Pan, Shirui},
  journal = {Nat. Mach. Intell.},
  volume = {7},
  pages = {437--447},
  year = {2025},
  doi = {10.1038/s42256-025-00994-z}
}

@article{zhuang2025advancing,
  title = {{Advancing Biomolecular Understanding and Design Following Human Instructions}},
  author = {Zhuang, Xiang and Ding, Keyan and Lyu, Tianwen and Jiang, Yinuo and Li, Xiaotong and Xiang, Zhuoyi and Wang, Zeyuan and Qin, Ming and Feng, Kehua and Wang, Jike and others},
  journal = {Nat. Mach. Intell.},
  volume = {7},
  pages = {1154--1167},
  year = {2025},
  doi = {10.1038/s42256-025-01064-0}
}

@article{liu2025quantitative,
  title = {{A Quantitative Analysis of Knowledge-Learning Preferences in Large Language Models in Molecular Science}},
  author = {Liu, Pengfei and Tao, Jun and Ren, Zhixiang},
  journal = {Nat. Mach. Intell.},
  volume = {7},
  pages = {315--327},
  year = {2025}
}

@article{liu2023multi,
  title = {{Multi-Modal Molecule Structure--Text Model for Text-Based Retrieval and Editing}},
  author = {Liu, Shengchao and Nie, Weili and Wang, Chengpeng and Lu, Jiarui and Qiao, Zhuoran and Liu, Ling and Tang, Jian and Xiao, Chaowei and Anandkumar, Animashree},
  journal = {Nat. Mach. Intell.},
  volume = {5},
  pages = {1447--1457},
  year = {2023}
}

@article{liu2025self,
  title = {{A Self-Feedback Knowledge Elicitation Approach for Chemical Reaction Predictions}},
  author = {Liu, Pengfei and Tao, Jun and Ren, Zhixiang},
  journal = {Eng. Appl. Artif. Intell.},
  volume = {156},
  pages = {111112},
  year = {2025}
}

@article{kuang2024impact,
  title = {{Impact of Domain Knowledge and Multi-Modality on Intelligent Molecular Property Prediction: A Systematic Survey}},
  author = {Kuang, Taojie and Liu, Pengfei and Ren, Zhixiang},
  journal = {Big Data Min. Anal.},
  volume = {7},
  pages = {858--888},
  year = {2024}
}

@article{wigh2022review,
  title = {{A Review of Molecular Representation in the Age of Machine Learning}},
  author = {Wigh, Daniel S and Goodman, Jonathan M and Lapkin, Alexei A},
  journal = {Wiley Interdiscip. Rev.: Comput. Mol. Sci.},
  volume = {12},
  pages = {e1603},
  year = {2022}
}

@article{li2025kolmogorov,
  title = {{Kolmogorov--Arnold Graph Neural Networks for Molecular Property Prediction}},
  author = {Li, Longlong and Zhang, Yipeng and Wang, Guanghui and Xia, Kelin},
  journal = {Nat. Mach. Intell.},
  volume = {7},
  pages = {1346--1354},
  year = {2025},
  doi = {10.1038/s42256-025-01087-7}
}

@inproceedings{vaswani2017attention,
  title = {{Attention Is All You Need}},
  author = {Vaswani, Ashish and Shazeer, Noam and Parmar, Niki and Uszkoreit, Jakob and Jones, Llion and Gomez, Aidan N and Kaiser, {\L}ukasz and Polosukhin, Illia},
  booktitle = {Advances in Neural Information Processing Systems 30},
  volume = {30},
  pages = {5998--6008},
  year = {2017}
}

@article{weininger1988smiles,
  title = {{{SMILES}, a Chemical Language and Information System. 1. Introduction to Methodology and Encoding Rules}},
  author = {Weininger, David},
  journal = {J. Chem. Inf. Comput. Sci.},
  volume = {28},
  pages = {31--36},
  year = {1988}
}

@inproceedings{devlin2019bert,
  title = {{{BERT}: Pre-Training of Deep Bidirectional Transformers for Language Understanding}},
  author = {Devlin, Jacob and Chang, Ming-Wei and Lee, Kenton and Toutanova, Kristina},
  booktitle = {Proceedings of the 2019 Conference of the North American Chapter of the Association for Computational Linguistics: Human Language Technologies},
  pages = {4171--4186},
  year = {2019}
}

@techreport{radford2019language,
  title = {{Language Models Are Unsupervised Multitask Learners}},
  author = {Radford, Alec and Wu, Jeffrey and Child, Rewon and Luan, David and Amodei, Dario and Sutskever, Ilya},
  institution = {OpenAI},
  type = {Technical Report},
  year = {2019},
  url = {https://cdn.openai.com/better-language-models/language_models_are_unsupervised_multitask_learners.pdf},
  note = {accessed 2026-08-03}
}

@article{raffel2020exploring,
  title = {{Exploring the Limits of Transfer Learning with a Unified Text-to-Text Transformer}},
  author = {Raffel, Colin and Shazeer, Noam and Roberts, Adam and Lee, Katherine and Narang, Sharan and Matena, Michael and Zhou, Yanqi and Li, Wei and Liu, Peter J},
  journal = {J. Mach. Learn. Res.},
  volume = {21},
  pages = {1--67},
  year = {2020}
}

@misc{schulman2022chatgpt,
  title = {{{ChatGPT}: Optimizing Language Models for Dialogue}},
  author = {{OpenAI}},
  year = {2022},
  url = {https://openai.com/index/chatgpt/},
  howpublished = {OpenAI},
  note = {accessed 2026-08-03}
}

@inproceedings{edwards2022translation,
  title = {{Translation between Molecules and Natural Language}},
  author = {Edwards, Carl and Lai, Tuan and Ros, Kevin and Honke, Garrett and Cho, Kyunghyun and Ji, Heng},
  booktitle = {Proceedings of the 2022 Conference on Empirical Methods in Natural Language Processing ({EMNLP})},
  pages = {375--413},
  year = {2022}
}

@inproceedings{pei2024biot5+,
  title = {{{BioT5+}: Towards Generalized Biological Understanding with {IUPAC} Integration and Multi-Task Tuning}},
  author = {Pei, Qizhi and Wu, Lijun and Gao, Kaiyuan and Liang, Xiaozhuan and Fang, Yin and Zhu, Jinhua and Xie, Shufang and Qin, Tao and Yan, Rui},
  booktitle = {Findings of the Association for Computational Linguistics: {ACL} 2024},
  pages = {1216--1240},
  year = {2024}
}

@book{favre2013nomenclature,
  title = {{Nomenclature of Organic Chemistry: {IUPAC} Recommendations and Preferred Names 2013}},
  author = {Favre, Henri A and Powell, Warren H},
  year = {2014},
  publisher = {Royal Society of Chemistry},
  address = {Cambridge},
  doi = {10.1039/9781849733069}
}

@article{liu2024git,
  title = {{{GIT-Mol}: A Multi-Modal Large Language Model for Molecular Science with Graph, Image, and Text}},
  author = {Liu, Pengfei and Ren, Yiming and Tao, Jun and Ren, Zhixiang},
  journal = {Comput. Biol. Med.},
  volume = {171},
  pages = {108073},
  year = {2024}
}

@inproceedings{yullasmol,
  title = {{{LLaSMol}: Advancing Large Language Models for Chemistry with a Large-Scale, Comprehensive, High-Quality Instruction Tuning Dataset}},
  author = {Yu, Botao and Baker, Frazier N and Chen, Ziqi and Ning, Xia and Sun, Huan},
  booktitle = {Proceedings of the First Conference on Language Modeling ({COLM})},
  year = {2024}
}

@misc{zhang2024chemllm,
  title = {{{ChemLLM}: A Chemical Large Language Model}},
  author = {Zhang, Di and Liu, Wei and Tan, Qian and Chen, Jingdan and Yan, Hang and Yan, Yuliang and Li, Jiatong and Huang, Weiran and Yue, Xiangyu and Ouyang, Wanli and Zhou, Dongzhan and Zhang, Shufei and Su, Mao and Zhong, Han-Sen and Li, Yuqiang},
  year = {2024},
  eprint = {2402.06852},
  archiveprefix = {arXiv},
  primaryclass = {cs.AI},
  doi = {10.48550/arXiv.2402.06852},
  url = {https://arxiv.org/abs/2402.06852},
  note = {submitted 2024-02-10; accessed 2026-08-03}
}

@misc{wang2025txgemma,
  title = {{{TxGemma}: Efficient and Agentic {LLMs} for Therapeutics}},
  author = {Wang, Eric and Schmidgall, Samuel and Jaeger, Paul F. and Zhang, Fan and Pilgrim, Rory and Matias, Yossi and Barral, Joelle and Fleet, David and Azizi, Shekoofeh},
  year = {2025},
  eprint = {2504.06196},
  archiveprefix = {arXiv},
  primaryclass = {cs.AI},
  doi = {10.48550/arXiv.2504.06196},
  url = {https://arxiv.org/abs/2504.06196},
  note = {submitted 2025-04-08; accessed 2026-08-03}
}

@article{wei2022chain,
  title = {{Chain-of-Thought Prompting Elicits Reasoning in Large Language Models}},
  author = {Wei, Jason and Wang, Xuezhi and Schuurmans, Dale and Bosma, Maarten and Ichter, Brian and Xia, Fei and Chi, Ed and Le, Quoc V and Zhou, Denny},
  journal = {Adv. Neural Inf. Process. Syst.},
  volume = {35},
  pages = {24824--24837},
  year = {2022}
}

@misc{openai2024openaio1card,
  title = {{{OpenAI o1} System Card}},
  author = {Jaech, Aaron and Kalai, Adam and Lerer, Adam and Richardson, Adam and El-Kishky, Ahmed and Low, Aiden and Helyar, Alec and Madry, Aleksander and Beutel, Alex and Carney, Alex and others},
  year = {2024},
  eprint = {2412.16720},
  archiveprefix = {arXiv},
  primaryclass = {cs.AI},
  doi = {10.48550/arXiv.2412.16720},
  url = {https://arxiv.org/abs/2412.16720},
  note = {submitted 2024-12-21; accessed 2026-08-03}
}

@misc{deepseekai2025deepseekr1,
  title = {{{DeepSeek-R1}: Incentivizing Reasoning Capability in {LLMs} via Reinforcement Learning}},
  author = {Guo, Daya and Yang, Dejian and Zhang, Haowei and Song, Junxiao and Zhang, Ruoyu and Xu, Runxin and Zhu, Qihao and Ma, Shirong and Wang, Peiyi and Bi, Xiao and others},
  year = {2025},
  eprint = {2501.12948},
  archiveprefix = {arXiv},
  primaryclass = {cs.CL},
  doi = {10.48550/arXiv.2501.12948},
  url = {https://arxiv.org/abs/2501.12948},
  note = {submitted 2025-01-22; accessed 2026-08-03}
}

@inproceedings{shazeer2017outrageously,
  title = {{Outrageously Large Neural Networks: The Sparsely-Gated Mixture-of-Experts Layer}},
  author = {Shazeer, Noam and Mirhoseini, Azalia and Maziarz, Krzysztof and Davis, Andy and Le, Quoc and Hinton, Geoffrey and Dean, Jeff},
  booktitle = {Proceedings of the 5th International Conference on Learning Representations ({ICLR})},
  year = {2017}
}

@article{wu2018moleculenet,
  title = {{{MoleculeNet}: A Benchmark for Molecular Machine Learning}},
  author = {Wu, Zhenqin and Ramsundar, Bharath and Feinberg, Evan N and Gomes, Joseph and Geniesse, Caleb and Pappu, Aneesh S and Leswing, Karl and Pande, Vijay},
  journal = {Chem. Sci.},
  volume = {9},
  pages = {513--530},
  year = {2018}
}

@inproceedings{banerjee2005meteor,
  title = {{{METEOR}: An Automatic Metric for {MT} Evaluation with Improved Correlation with Human Judgments}},
  author = {Banerjee, Satanjeev and Lavie, Alon},
  booktitle = {Proceedings of the {ACL} Workshop on Intrinsic and Extrinsic Evaluation Measures for Machine Translation and/or Summarization},
  pages = {65--72},
  year = {2005}
}

@article{durant2002reoptimization,
  title = {{Reoptimization of {MDL} Keys for Use in Drug Discovery}},
  author = {Durant, Joseph L and Leland, Burton A and Henry, Douglas R and Nourse, James G},
  journal = {J. Chem. Inf. Comput. Sci.},
  volume = {42},
  pages = {1273--1280},
  year = {2002}
}

@misc{liu2024deepseek,
  title = {{{DeepSeek-V3} Technical Report}},
  author = {Liu, Aixin and Feng, Bei and Xue, Bing and Wang, Bingxuan and Wu, Bochao and Lu, Chengda and Zhao, Chenggang and Deng, Chengqi and Zhang, Chenyu and Ruan, Chong and others},
  year = {2024},
  eprint = {2412.19437},
  archiveprefix = {arXiv},
  primaryclass = {cs.CL},
  doi = {10.48550/arXiv.2412.19437},
  url = {https://arxiv.org/abs/2412.19437},
  note = {submitted 2024-12-27; accessed 2026-08-03}
}

@misc{comanici2025gemini,
  title = {{{Gemini 2.5}: Pushing the Frontier with Advanced Reasoning, Multimodality, Long Context, and Next Generation Agentic Capabilities}},
  author = {Comanici, Gheorghe and Bieber, Eric and Schaekermann, Mike and Pasupat, Ice and Sachdeva, Noveen and Dhillon, Inderjit and Blistein, Marcel and Ram, Ori and Zhang, Dan and Rosen, Evan and others},
  year = {2025},
  eprint = {2507.06261},
  archiveprefix = {arXiv},
  primaryclass = {cs.CL},
  doi = {10.48550/arXiv.2507.06261},
  url = {https://arxiv.org/abs/2507.06261},
  note = {submitted 2025-07-07; accessed 2026-08-03}
}

@techreport{anthropic2025claude35,
  title = {{{Claude 3.5 Sonnet} Model Card Addendum}},
  author = {{Anthropic}},
  institution = {Anthropic},
  type = {Model Card Addendum},
  year = {2024},
  url = {https://www-cdn.anthropic.com/fed9cc193a14b84131812372d8d5857f8f304c52/Model_Card_Claude_3_Addendum.pdf},
  note = {accessed 2026-08-03}
}

@techreport{openai2025o4mini,
  title = {{{OpenAI o3 and o4-mini} System Card}},
  author = {{OpenAI}},
  institution = {OpenAI},
  type = {System Card},
  year = {2025},
  url = {https://cdn.openai.com/pdf/2221c875-02dc-4789-800b-e7758f3722c1/o3-and-o4-mini-system-card.pdf},
  note = {accessed 2026-08-03}
}

@misc{yang2025qwen3,
  title = {{{Qwen3} Technical Report}},
  author = {Yang, An and Li, Anfeng and Yang, Baosong and Zhang, Beichen and Hui, Binyuan and Zheng, Bo and Yu, Bowen and Gao, Chang and Huang, Chengen and Lv, Chenxu and others},
  year = {2025},
  eprint = {2505.09388},
  archiveprefix = {arXiv},
  primaryclass = {cs.CL},
  doi = {10.48550/arXiv.2505.09388},
  url = {https://arxiv.org/abs/2505.09388},
  note = {submitted 2025-05-14; accessed 2026-08-03}
}

@misc{mistral2025ministrale8b,
  title = {{Un Ministral, des Ministraux}},
  author = {{Mistral AI}},
  year = {2024},
  url = {https://mistral.ai/news/ministraux},
  howpublished = {Mistral AI},
  note = {accessed 2026-08-03}
}

@misc{meta2025llama4,
  title = {{The {Llama 4} Herd: The Beginning of a New Era of Natively Multimodal {AI} Innovation}},
  author = {{Meta AI}},
  year = {2025},
  url = {https://ai.meta.com/blog/llama-4-multimodal-intelligence/},
  howpublished = {Meta AI},
  note = {accessed 2026-08-03}
}

@inproceedings{hu2022lora,
  title = {{LoRA: Low-Rank Adaptation of Large Language Models}},
  author = {Hu, Edward J. and Shen, Yelong and Wallis, Phillip and Allen-Zhu, Zeyuan and Li, Yuanzhi and Wang, Shean and Wang, Lu and Chen, Weizhu},
  booktitle = {Proceedings of the 10th International Conference on Learning Representations ({ICLR})},
  year = {2022}
}

@article{mcinnes2018umap,
  title = {{{UMAP}: Uniform Manifold Approximation and Projection}},
  author = {McInnes, Leland and Healy, John and Saul, Nathaniel and Gro{\ss}berger, Lukas},
  journal = {J. Open Source Softw.},
  volume = {3},
  pages = {861},
  year = {2018},
  doi = {10.21105/joss.00861}
}

@misc{lu2024aiscientistfullyautomated,
  title = {{The {AI} Scientist: Towards Fully Automated Open-Ended Scientific Discovery}},
  author = {Lu, Chris and Lu, Cong and Lange, Robert Tjarko and Foerster, Jakob and Clune, Jeff and Ha, David},
  year = {2024},
  eprint = {2408.06292},
  archiveprefix = {arXiv},
  primaryclass = {cs.AI},
  doi = {10.48550/arXiv.2408.06292},
  url = {https://arxiv.org/abs/2408.06292},
  note = {submitted 2024-08-12; accessed 2026-08-03}
}

@misc{gottweis2025aicoscientist,
  title = {{Accelerating Scientific Discovery with {Co-Scientist}}},
  author = {Gottweis, Juraj and Weng, Wei-Hung and Daryin, Alexander and Tu, Tao and Sirkovic, Petar and Myaskovsky, Artiom and Glowaty, Grzegorz and Weissenberger, Felix and Orlandi, Alessio and Popovici, Dan and others},
  year = {2025},
  eprint = {2502.18864},
  archiveprefix = {arXiv},
  primaryclass = {cs.AI},
  doi = {10.48550/arXiv.2502.18864},
  url = {https://arxiv.org/abs/2502.18864},
  note = {submitted 2025-02-26; accessed 2026-08-03}
}

@article{van2008visualizing,
  title = {{Visualizing Data Using {t-SNE}}},
  author = {Van der Maaten, Laurens and Hinton, Geoffrey},
  journal = {J. Mach. Learn. Res.},
  volume = {9},
  pages = {2579--2605},
  year = {2008}
}

@article{paul2024muscarinic,
  title = {{Muscarinic Receptor Activators as Novel Treatments for Schizophrenia}},
  author = {Paul, Steven M and Yohn, Samantha E and Brannan, Stephen K and Neugebauer, Nichole M and Breier, Alan},
  journal = {Biol. Psychiatry},
  volume = {96},
  pages = {627--637},
  year = {2024}
}

\end{document}